\documentclass[conference]{IEEEtran}
\IEEEoverridecommandlockouts
\usepackage{cite}
\usepackage{amsmath,amssymb,amsfonts}
\usepackage{algorithmic}
\usepackage{graphicx}
\usepackage{textcomp}
\usepackage{xcolor}
\usepackage{newtxmath}

\usepackage{bm}
\usepackage{bbm}
\usepackage{booktabs}
\usepackage{multirow}
\usepackage{algorithmic}
\usepackage{mathtools}
\usepackage{array,tabularx}
\usepackage{wrapfig}

\newcommand{\etal}{et al.}

\usepackage[ruled,vlined, linesnumbered]{algorithm2e}
\SetKwInput{KwInput}{Input}                
\SetKwInput{KwOutput}{Output}              
\SetKwComment{Comment}{$\triangleright$\ }{}
\usepackage[switch]{lineno}

\def\BibTeX{{\rm B\kern-.05em{\sc i\kern-.025em b}\kern-.08em
    T\kern-.1667em\lower.7ex\hbox{E}\kern-.125emX}}

\usepackage[breaklinks=true,colorlinks,bookmarks=false]{hyperref}
    
\begin{document}

\title{Smooth Variational Graph Embeddings for Efficient Neural Architecture Search
\thanks{The authors acknowledge support by the German Federal Ministry of Education and Research Foundation via the project DeToL.}
}
\author{Jovita Lukasik\textsuperscript{\rm 1},
        David Friede\textsuperscript{\rm 1},
        Arber Zela\textsuperscript{\rm 2}, 
        Frank Hutter\textsuperscript{\rm 2,}\textsuperscript{\rm 3},
        Margret Keuper\textsuperscript{\rm 1}\\
    \textsuperscript{\rm 1}University of Mannheim, 
    \textsuperscript{\rm 2}University of Freiburg,
    \textsuperscript{\rm 3}Bosch Center for Artificial Intelligence\\
    {\tt\small \{jovita, david\}@informatik.uni-mannheim.de, keuper@uni-mannheim.de}\\
    {\tt\small \{zelaa, fh\}@cs.uni-freiburg.de} \\
    }

\maketitle

\begin{abstract}
Neural architecture search (NAS) has recently been addressed from various directions, including discrete, sampling-based methods and efficient differentiable approaches. While the former are notoriously expensive, the latter suffer from imposing strong constraints on the search space. Architecture optimization from a learned embedding space for example through graph neural network based variational autoencoders builds a middle ground and leverages advantages from both sides. Such approaches have recently shown good performance on several benchmarks. Yet, their stability and predictive power heavily depends on their capacity to reconstruct networks from the embedding space. In this paper, we propose a two-sided variational graph autoencoder, which allows to smoothly encode and accurately reconstruct neural architectures from various search spaces. We evaluate the proposed approach on neural architectures defined by the ENAS approach, the NAS-Bench-101 and the NAS-Bench-201 search space and show that our smooth embedding space allows to directly extrapolate the performance prediction to architectures outside the seen domain (e.g. with more operations). 
Thus, it facilitates to predict good network architectures even without expensive Bayesian optimization or reinforcement learning. 

 \end{abstract}


\section{Introduction}
Recent progress in computer vision is to a large extent coupled to the advancement of novel neural architectures \cite{krizhevsky2012imagenet,goodfellow2014generative}. In this context, the automated search of neural architectures \cite{Real2017,zoph2018learning,Real2019}
is increasingly important, as it removes the fatiguing and time-consuming process of manual trial-and-error network design.
 \begin{figure}[t]
    \centering
    \includegraphics[width=\columnwidth]{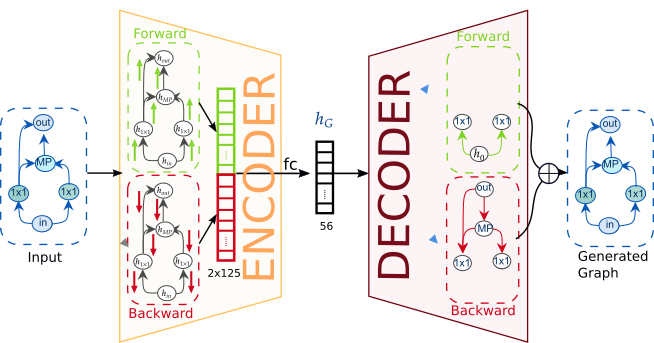}
    \caption{Architecture of the proposed SVGe model.
    It takes as input a neural architecture graph. 
    The encoder (left) uses two GNN modules, the forward encoder (green) and the backward encoder (red), to create an informative latent representation $\mathbf{h}_G$. This latent vector is input to the decoder (right), which decodes forward (green) and backward (red) pass separately, generating two graphs in a sequential manner. Their union is the output of SVGe. }
    \label{fig:SVGe}
\end{figure}
Neural Architecture Search (NAS) is intrinsically a discrete optimization problem and can be solved effectively using black-box methods such as  
reinforcement learning \cite{Zoph2017,zoph2018learning}, evolution \cite{elsken2018neural,Real2019}, Bayesian optimization (BO) \cite{Kandasamy2018,white2019bananas, Ru2020NeuralAS} or local search \cite{white2020local}. However, finding a good solution typically requires thousands of function evaluations, which is infeasible without company-scale compute infrastructure. Recent research in NAS focus as well on efficient methods via continuous relaxations of the discrete search space and weight-sharing \cite{bender_icml:2018, Pham2018, Liu2018, Cai19}. However, such methods
yield efficient yet oftentimes sub-optimal results~\cite{zela19}.

Therefore, we argue in favor of NAS on learned graph embeddings using encoder-decoder graph neural networks (GNN)~\cite{gori2005new,kipf2016semi,wu2019comprehensive}. Zhang \etal~\cite{zhang2019d} recently showed good performance with such a model, D-VAE, on the ENAS search space~\cite{Pham2018} in neural architecture performance prediction and BO - proving its ability to learn smooth continuous graph representations. D-VAE aggregates information in the architecture GNN alternatingly in the forward pass and in the backward pass to encode the neural network information flow. However, the D-VAE model imposes strong constraints on the graph structure, which limit its applicability to search spaces beyond ENAS. In addition, it has very long training times. 
In this paper, we propose a two-sided variational encoder-decoder GNN to learn smooth embeddings in various NAS search spaces, which we call \textit{Smooth Variational Graph embedding} (SVGe). In contrast to D-VAE, SVGe aggregates node representations in the forward and backward pass separately and consequently decodes their joint representation into forward and backward pass separately (see Fig. \ref{fig:SVGe}). This yields a very high reconstruction ability without imposing any constraints on the search space and allows for a more efficient training. 

Inheriting from variational autoencoders \cite{kingma2013auto}, it places structurally similar graphs close to one another in the embedding space and thus facilitates efficient black-box optimization to find high-performing architectures. The proposed model is not only three times faster than D-VAE but also shows improved BO results on the ENAS search space. In contrast to D-VAE, it can be directly applied to other search spaces such as NAS-Bench-101~\cite{ying2019bench} and NAS-Bench-201~\cite{dong2020}.

Moreover, it allows to learn architecture performance prediction in a supervised way and extrapolate from the space of observed architectures at test time. This way, high performing architectures even outside of the original search space can be proposed at very low costs.

In summary, we make the following contributions:
    (i) We introduce a novel graph variational autoencoder, SVGe, that builds a structurally smooth variational graph embedding by learning accurate representations of neural architectures (Sec.~\ref{subsec:encoder} and \ref{sec:decoder}). 
    (ii) We discuss theoretical properties of our approach 
    (Sec.~\ref{sec:discussion}). 
    (iii) We conduct extensive evaluations on the ENAS~\cite{Pham2018}, NAS-Bench-101~\cite{ying2019bench} and NAS-Bench-201\cite{dong2020} search spaces and show that our approach allows for competitive BO results in all three search spaces. Our experiments show that SVGe is able to extrapolate to larger unseen architectures. It finds an architecture with a best accuracy of $95.18 \%$ when learning from the NAS-Bench-101 search space. This improves over the best architecture within this space. In addition, our top $1$ found architecture improves over comparable architectures in terms of validation and test accuracy, when transferring to ImageNet16-120 \cite{ImageNet16_120}      
    (Sec.~\ref{sec:experiments}). 
    
 
\section{Related Work}\label{sec:related}

\paragraph{Graph Generative Models.}
Recent years have shown huge progress in representation learning for graph-based data with Graph Neural Networks (GNNs)~\cite{li2016, kipf2016semi, niepert2016learning, hamilton2017inductive}. GNNs follow a message passing scheme, where node feature vectors aggregate information from their neighbors~\cite{gilmer2017neural} and capture local structural information. 
To obtain a graph-level representation these 
feature vectors are pooled \cite{ying2018}.
GNNs differ in their neighborhood node information as well as in their graph-level aggregation procedure 
\cite{scarselli2009,hamilton2017inductive,kipf2016semi, li2016, Velickovic18,xu2018, GIN}.
Graph generation can be addressed globally by relaxing the adjacency matrix \cite{kipf2016variational, simonovsky2018graphvae} or sequentially by adding nodes and edges alternately using recurrent networks \cite{luo2018neural,you2018graphrnn} or GNNs 
\cite{li2018learning}. Our decoder model is similar to \cite{li2018learning} in the aggregation procedure. Yet, while \cite{li2018learning} treat forward and backward pass equally, our model aggregates node information for both separately to account for the order of network operations and the information flow. Zhang \etal~\cite{zhang2019d} propose a less efficient, alternating message passing scheme for this purpose and reinstall the validity of decoded architectures using a heuristic which employs prior knowledge on the search space. The proposed method differs in both encoder and decoder. Our encoder employs an efficient bi-directional model and the proposed bi-directional decoding facilitates highly accurate reconstructions without constraining the search space.

\paragraph{Neural Network Performance Prediction.}
Predicting the performance of neural networks based on features such as the network architecture, training hyperparameters or learning curves has been exploited previously via MCMC methods~\cite{domhan-ijcai15}, Bayesian neural networks~\cite{klein-iclr17} or regression models~\cite{baker-iclr17}. \cite{white2019bananas} and \cite{Long2019PerformancePB} manually construct features to regress neural networks or support vector regressors.

More recent work utilizes GNN encodings by adapting message passing to simulate operations in either edges or nodes in the graph~\cite{ning_gates} or using a semi-supervised approach by training GNNs on relation graphs in the latent space~\cite{Tang_Assessor}. 
\cite{Lukasik2020} 
train a GNN surrogate performance prediction model and show the ability of zero-shot performance prediction.  

\paragraph{Neural Architecture Search via Bayesian Optimization.}
To apply Bayesian Optimization (BO) for NAS, high dimensional and discrete architecture configurations need to be embedded into continuous search spaces. 
\cite{Kandasamy2018} use a distance metric obtained through an optimal transport program to enable Gaussian process (GP)-based BO. \cite{white2019bananas} encode architectures with a high-dimensional path-based scheme and employ BO on an ensemble surrogate. \cite{Ru2020NeuralAS} propose a graph kernel with GP-based BO to capture the topological structure of architecture graphs. GNN-based encodings have been used in~\cite{shi2019multi,zhang2019d} to fit a Bayesian linear regressor as a surrogate in BO.
Very recently, Yan~\etal~\cite{Arch2vec} learn neural architecture representations using \cite{GIN} in combination with a multilayer perceptron graph decoder. The model employs a combination of the adjacency matrix and a one-hot operation encoding matrix as input for the encoder and improves over previous approaches to NAS though. Their results indicate that highly informed encoding is crucial for the task.
\color{black}
Our model focuses on learning an accurate architecture mapping into a smooth latent space using GNNs. It allows competitive performance to highly optimized approaches using BO \cite{Arch2vec}. Due to its smoothness, it can further directly extrapolate the performance prediction outside the space of seen architectures and thus to propose high-performing, deeper networks without using BO.

\section{Structural Graph Autoencoding}
We aim to learn a structurally smooth latent representation of neural network architectures, which we cast as directed acyclic graphs (DAGs) with nodes representing operations (like convolution or pooling) and edges representing information flow. 
This enables to 
(i) accurately predict the accuracy of an unseen graph from training samples and (ii) draw new samples which are structurally similar to previously seen ones. 

Our model is a \textit{variational autoencoder (VAE)}~\cite{kingma2013auto}. First, the VAE encoder $q_{\phi}(\textbf{z} \vert G)$ maps the input data $G$ (a finite number of i.i.d. samples from an unknown distribution) onto a continuous latent variable $\textbf{z}$ via a parametric function $q_{\phi}$. Then a probabilistic generative model $p_{\theta}(G \vert \textbf{z})$ (the decoder) decodes the latent variables $\textbf{z}$ to the original representation. The parameters $\phi$ and $\theta$ of the encoder and decoder are optimized by maximizing the evidence lower bound (ELBO):
\begin{equation}\label{VAE}
{\mathcal{L}}(\theta, \phi ; G) = \mathbb{E}_{q_{\phi}(\mathbf{z} \vert  G)} \big[\log p_{\theta}(G \vert \mathbf{z}) \big] 
 - \text{D}_{\text{KL}} \left(q_{\phi}(\mathbf{z} \vert G) \Vert p(\mathbf{z})\right)
\end{equation}
where the first term is the reconstruction loss and enforces high similarity between input data and generated data, while the second term is the Kullback-Leibler divergence which regularizes the latent space. 
From the trained VAE, new data can be generated by decoding latent space variables $\textbf{z}$ sampled from the prior distribution $p(\mathbf{z})$.

Below, we provide details on the proposed GNN encoder and decoder models. For NAS, we have to pay particular attention to isomorphic graphs. As they are functionally identical yet represented via distinct adjacency matrices, it is not obvious to guarantee a correct mapping and unique decoding. Motivated by \cite{GIN} (see Sec.~\ref{sec:discussion}), we chose our model to allow for injective encoding and unique decoding.

\subsection{Encoder} \label{subsec:encoder}
Here, we describe the encoder of our GNN-based SVGe. 
Let $G=(V, E)$ be a directed acyclic graph, with nodes $\mathrm{v}\in V$ and
edges $\mathrm{e} \in E$. Each node $\mathrm{v}$ has an initial node feature embedding $\mathbf{h}^{(0)}_{\mathrm{v}}$. 
Standard GNNs can be seen as a two-step procedure. In the first step the GNN learns a representation for each node $\mathrm{v} \in V$, by iteratively aggregating the representations of neighboring nodes using an aggregation function $\mathcal{A}(\cdot)$. Then it updates the representation with the update function $\mathcal{U}(\cdot)$. After $K$ rounds of iterations, the final representation of each node $\mathrm{v}$ is computed. 
%
The second step computes a graph representation $\mathbf{h}_G$ by aggregating 
the node representations. 

Since our objective is to learn a structurally smooth graph representation, 
we need to capture the structure and the information flows of the graphs. 
Thus, our model consists of two encoding modules, where the messages are passed in the direction of the network's forward pass in the forward encoder (green) and in the direction of the back-propagation in the backward encoder (red) visualized in Fig.~\ref{fig:SVGe} (left).  
Our variant of GNN formulates the aggregation function $\mathcal{A}(\cdot)$ as the sum of node message passing modules and uses a single gated recurrent unit (GRU)~\cite{gru} as the update function $\mathcal{U}(\cdot)$ for both forward and backward encoder. 
The forward message passing module $\overrightarrow{f}\big(\overrightarrow{\mathbf{h}}{}^{(k)}_{\mathrm{u}}, \overrightarrow{\mathbf{h}}{}^{(k)}_{\mathrm{v}}\big)$
computes a message vector from node $\mathrm{u}$ to node $\mathrm{v}$ in the $k$-th iteration, while 
$\overleftarrow{f}\big(\overleftarrow{\mathbf{h}}{}^{(k)}_{\mathrm{v}}, \overleftarrow{\mathbf{h}}{}^{(k)}_{\mathrm{u}}\big)$ 
is the backward message passing module from node $\mathrm{v}$ to $\mathrm{u}$, with  $\mathbf{h}_{\mathrm{v}}^{(k)}$ being a feature vector representation of node $\mathrm{v}$ at iteration $k$.
Each graph information direction is aggregated individually:
\begin{equation}\label{eq:GNN_node_aggregation}
\begin{split}
 \overrightarrow{\mathbf{a}}_{\mathrm{v}}{}^{(k)} &=\sum_{ \mathrm{u} \in \mathcal{V}^{\mathrm{in}}(\mathrm{v})}
 \overrightarrow{f}\Big(\overrightarrow{\mathbf{h}}{}^{(k-1)}_{\mathrm{u}}, \overrightarrow{\mathbf{h}}{}^{(k-1)}_{\mathrm{v}}\Big)  \\
 \overleftarrow{\mathbf{a}}_{\mathrm{v}}{}^{(k)} &=\sum_{ \mathrm{u} \in \mathcal{V}^{\mathrm{out}}(\mathrm{v})}  
 \overleftarrow{f}\Big(\overleftarrow{\mathbf{h}}{}^{(k-1)}_{\mathrm{v}}, \overleftarrow{\mathbf{h}}{}^{(k-1)}_{\mathrm{u}}\Big) ,
    \end{split}
\end{equation}
where $\mathcal{V}^{\mathrm{in}}(\mathrm{v})=\{ \mathrm{u} \in V \mid (\mathrm{u},\mathrm{v}) \in E\} $ is the set of adjacent nodes to $\mathrm{v}$ in the DAG, specifying the network input to $\mathrm{v}$ during inference and $\mathcal{V}^{\mathrm{out}}(\mathrm{v})= \{ \mathrm{u} \in V \mid (\mathrm{v},\mathrm{u}) \in E\}$ are the adjacent nodes to $\mathrm{v}$ in the networks backward pass, i.e. in the DAG with reversed edges.
Also, instead of using one-hot encoded node labels, we employ a learnable embedding table $\mathrm{L_e}$ on the node types, which stores embeddings (feature vectors) for our initial node embeddings $\overrightarrow{\mathbf{h}}{}^{(0)}_{\mathrm{v}}, \overleftarrow{\mathbf{h}}{}^{(0)}_{\mathrm{v}}$. 
The functions $\overrightarrow f$ and $\overleftarrow f$ are implemented using a single fully connected layer (fc).
After the final iteration, we combine the forward and backward node embeddings $\big(\overrightarrow{\mathbf{h}}{}^{(K)}_{\mathrm{v}}\big)_{\mathrm{v} \in \mathcal{V}} $,  $\big(\overleftarrow{\mathbf{h}}{}^{(K)}_{\mathrm{v}}\big)_{\mathrm{v} \in \mathcal{V}}$ 
 of node $\mathrm{v}$ by concatenation:
\begin{equation}\label{eq:enc_node_concat}
  \Big(\mathbf{h}^{(K)}_{\mathrm{v}}\Big)_{\mathrm{v} \in V}  = \Big( \mathrm{CONCAT}\big(\overrightarrow{\mathbf{h}}{}^{(K)}_{\mathrm{v}}, \overleftarrow{\mathbf{h}}{}^{(K)}_{\mathrm{v}}\big)\Big )_{\mathrm{v} \in V}.
\end{equation}
By combining these two information sets, we capture not only the topology but also the information paths in the graph.

From the node representations, we compute a graph representation using 
a gated sum: 
\begin{equation}\label{eq:GraphAggr}
    \mathbf{h}_G= \sum_{\mathrm{v} \in V} \phi \Big(\mathbf{h}_{\mathrm{v}}^{(K)}\Big) \odot \psi\Big(\mathbf{h}_{\mathrm{v}}^{(K)}\Big),
\end{equation}
with the sigmoid activated fc layer $\phi= \sigma(\mathrm{fc}(\cdot))$ as a gating function, the linear activated fc layer $\psi$
, and $\odot$ denoting the Hadamard product. 
Since we use this encoder in a variational autoencoder setting, we add an extra graph aggregation layer equal to \eqref{eq:GraphAggr} to obtain $\mathbf{h}_G^{\mathrm{var}}$. Thus, the outputs of our encoder are the parameters of the approximate posterior distribution $q_{\phi}= \mathcal{N}(\mathbf{h}_G, \bm{\Sigma})$, with $\mathbf{h}_G$ being the mean and $\mathbf{h}_G^{\mathrm{var}}$ the diagonal of the variance-covariance matrix $\bm{\Sigma}$ of the multivariate normal distribution. 
Sec. \ref{sec:discussion} discusses the properties of this encoder w.r.t. injectivity and isomorphic graphs. 

\subsection{Decoder}\label{sec:decoder}

The SVGe decoder $p_{\theta}(G \vert \textbf{z})$ takes a latent point $\textbf{z}$ 
as input and reconstructs $G$ simultaneously from two directions (start-to-end and end-to-start node), see Fig.~\ref{fig:SVGe} (right). As in the encoder, the model explicitly learns a neural architecture's forward \emph{and} backward pass, allowing for highly accurate reconstructions of graphs without ``loose ends". 
%
%
%
The directional graph generation starts from the input node $\mathrm{v_0}$ for forward decoding and the output node $\mathrm{v_T}$ for backward decoding. Each graph is built iteratively in a sequence of operations that add nodes and edges until the end node is generated, similar to \cite{zhang2019d}. 
%
The union of both, forward and backward graph, builds the output graph.

\subsubsection{Directional Decoding.} The directional graph generation starts from an initial node with type ``InputType" for the forward decoding and with type ``OutputType" for the backward decoding. 
The input node and all generated nodes $\mathrm{v_t}$ are embedded according to the \emph{initNode} module:
\begin{equation}\label{eq:initNode}
    \mathbf{h}_{t} = f_{\mathrm{initNode}}\big(\mathbf{z}, \mathbf{h}_{\widetilde{G}^{(t)}}, \mathrm{L_d}[\mathrm{type}]\big),
\end{equation}
with $f_{\mathrm{initNode}}$ being a two-layer fc with ReLU activation. It takes as input the sampled point $\mathbf{z}$, the partial graph embedding $\mathbf{h}_{\widetilde{G}^{(t)}}$ and the learned node type embedding $\mathrm{L_d}[\mathrm{type}]$. 
If the node $\mathrm{v_t}$ is either the input or the output node, Eq. \eqref{eq:initNode} simplifies to $\mathbf{h}_{t} = f_{\mathrm{initNode}}(\mathbf{z}, \mathrm{L_d}[\mathrm{type}])$.
%
Given the start node $\mathrm{v_0}$, its embedding $\mathbf{h}_0$ and the partial graph embedding $\mathbf{h}_G = \mathbf{z}$, a graph is generated by iterating over a sequence of modules whose 
weights are shared across iterations until an end node, e.g. of type OutputNode in the forward decoder, is drawn.

In every iteration, a new node is created and added to the graph and its node type (i.e. operation in the network architecture) is selected by the \emph{addNode} module. It takes as input the representation of the partial graph $\mathbf{h}_{\widetilde{G}^{(t)}}$ and the sampled point $\mathbf{z}$ and determines the next missing 
node
\begin{equation}\label{eq:node_type_module}
     \mathrm{NodeType} \sim \mathrm{Categorical}\Big(s_{\mathrm{addNode}}^{(t+1)}\Big),
\end{equation}
where 
\begin{equation} \label{eq:addNode}
    s_{\mathrm{addNode}}^{(t+1)} = f_{\mathrm{addNode}}\big(\mathbf{z}, \mathbf{h}_{\widetilde{G}^{(t)}}\big). \\ 
\end{equation}
$ f_{\mathrm{addNode}}$ are two fc layers with ReLU-non-linearities, producing
parameters for the categorical node type distribution from which the next node is sampled. 
This newly added node is then initialized with the  \textit{initNode} module (Eq. \eqref{eq:initNode}) 
%
and added to the already existing propagated node embedding, $\mathbf{h}_{\widetilde{V}}=\mathrm{CONCAT}((\mathbf{h}_{j})_{0\leq j \leq t},\mathbf{h}_{t+1})$.

To every added node $\mathrm{v_{t+1}}$, the decoder creates edges from already existing nodes according to a scoring function, where a high value represents a likely edge. 
This \emph{addEdges} module takes as input the partial graph node embeddings $\mathbf{h}_{t+1}$ and $\mathbf{h}_{t} = (\mathbf{h}_j)_{0 \leq j \leq t}
$, as well as the partial graph embedding $\mathbf{h}_{\widetilde{G}^{(t)}}$ and the sampled point $\mathbf{z}$, leading to
\begin{equation} \label{eq:add_edge_module}
    e_{(i,t+1)} \sim \mathrm{Ber}\big(f_{\mathrm{addEdges}}\big(\mathbf{h}_{t+1},\mathbf{h}_t, \mathbf{h}_{\widetilde{G}^{(t)}}, \mathbf{z}\big)\big),
\end{equation}
where $\mathrm{Ber}$ denotes a Bernoulli distribution. Sampling from this distribution yields the new set of directed edges ending in $\mathrm{v_{t+1}}$. 
$f_{\mathrm{addEdges}}$ is again a two-layer fc layer with ReLU activation.
%

After adding the edges, the concatenated node embeddings $\mathbf{h}_{\widetilde{V}}$ are aggregated and updated in the \emph{prop} module as in the encoder according to Eq. \eqref{eq:GNN_node_aggregation}, 
yielding the updated node embedding $\mathbf{h}$. These node embeddings are aggregated into a single graph representation $\mathbf{h}_G$ according to Eq.~\eqref{eq:GraphAggr}. 
Encoder and decoder GNNs have distinct weights. 

\subsection{Loss Function and Training} \label{sec:training}
%
We train the encoder and decoder of SVGe jointly in an unsupervised manner.
Given a fixed node ordering of the DAG, which we discuss in Sec. \ref{sec:discussion}, we know the ground truth of the outputs of \textit{AddNode} (Eq.~ \eqref{eq:node_type_module}) and \textit{AddEdges} (Eq.~\eqref{eq:add_edge_module}) during training.
We use this ground truth to compute a node-level loss $\mathcal{L}_V^t$ and an edge-level loss $\mathcal{L}_{E}^t$ at each iteration $t$. 
Additionally, we replace the model output by the ground truth such that possible errors will not accumulate throughout iterations. 
To compute the overall reconstruction loss for a graph $G$, we sum up node losses and edge losses over all iterations for both decoding directions;
$\mathcal{L}_{\mathrm{rec}} = \overrightarrow{\mathcal{L}_V} + \overrightarrow{\mathcal{L}_{E}} + \overleftarrow{\mathcal{L}_V} + \overleftarrow{\mathcal{L}_{E}}.$
Following \cite{kingma2013auto}, we assume $p_{\theta}(\textbf{z})~\sim~\mathcal{N}(\textbf{z};\mathbf{0}, \mathbbm{1})$ and $p_{\theta}(G \vert \textbf{z}) \sim \mathcal{N}( \mathbf{h}_G, \bm{\Sigma})$. Furthermore, we approximate the posterior by a multivariate Gaussian distribution with diagonal covariance structure. This can be written as $\log q_{\phi}(\textbf{z} \vert G) = \log \mathcal{N}(\textbf{z};\mathbf{h}_G, \bm{\Sigma})$ and ensures a closed form of the  $\mathrm{KL}$ divergence
\begin{equation} 
\text{D}_{\text{KL}}=- \frac{1}{2} \sum_{j=1}^{J} \big(1+ \log (\mathbf{h}^{\mathrm{var}}_{G})_j - (\mathbf{h}_{G})_j^2 - (\mathbf{h}^{\mathrm{var}}_{G})_j\big).
\end{equation}
Thus, the overall loss from Eq. \eqref{VAE} becomes:
 \begin{equation} \label{SVGe_loss} 
{\mathcal{L}} =
\mathcal{L}_{\mathrm{rec}}+ \alpha\text{D}_{\text{KL}}.
\end{equation}
Following \cite{Jin2018} and \cite{zhang2019d}, we set $\alpha=0.005$.

\subsection{Discussion on Theoretical Properties} \label{sec:discussion}
In the following, we discuss SVGe in the context of isomorphic input graphs. Intuitively, if isomorphic graphs, i.e. graph representations of the same neural architecture, are mapped to distinct latent points, the latent space is intrinsically redundant. This hampers an efficient embedding of structural similarity. Conversely, if non-isomorphic graphs are mapped to the same latent point, their performance can neither be correctly predicted nor their structure reconstructed. Thus, a suitable graph encoder has to map any two isomorphic graphs to the same latent point (\textit{ISO}). A suitable decoder decodes each latent point to a unique graph and preserves the difference between non-isomorphic graphs (\textit{INJ}).  

\paragraph{Unique Latent Space Representation.}\label{enc_injectiv}
Here, we discuss the proposed SVGe \emph{encoder} w.r.t. properties (\textit{ISO}) and (\textit{INJ}).
\textit{Theorem 3} in \cite{GIN} gives sufficient conditions on injectivity (\textit{INJ}) of the GNN's node aggregation module and its update module. However, the required existence of an appropriate injective aggregation function operating on multisets can only be guaranteed theoretically on countable input feature spaces. Even then, \cite{GIN} gives no explicit construction but they argue that it can be approximated via MLPs. 
Thus, we verify empirically that SVGe maps isomorphic graphs to the same embedding (\textit{ISO}) in an experiment on $11,606$ isomorphic graph pairs of length $7$ from the NAS-Bench-101 search space. 100\% of such isomorphic pairs were mapped onto the same point. 
The mapping of distinct graphs to distinct latent representations (\textit{INJ}) is a prerequisite for accurate reconstruction and therefore validated in the reconstruction ability in Sec.~\ref{sec:exp_abilities}. 

\paragraph{Decoding from the Latent Space.}
We now discuss how the \emph{decoder} handles isomorphic DAGs w.r.t.~(\textit{INJ}). Since isomorphic graphs are mapped onto the same latent point by the encoder (see above), it suffices for the decoder to decode them uniquely, i.e. deterministically. It can be easily seen that this is the case for SVGe. 
Yet, how can the decoder be trained efficiently to decode one out of several isomorphic graphs from the same latent point, when decoding graph $G_1$ instead of an isomorphic graph $G_2$ leads to significant reconstruction loss? Isomorphic graphs can be created from one another by permuting nodes in the adjacency matrix. Thus, to ease the decoder into the learning process 
we bring the graphs in a unified form. 
Towards a unique representation, we limit the training data to upper triangular matrices.
%
The remaining isomorphic graphs are removed from the training set as in \cite{ying2019bench}, since we need a unique representation for each graph. 
Note, this only removes duplicate architectures.
For the ENAS and the NAS-Bench-201 search space, the adjacency matrices are unique, given the upper triangular representation. 

Given such clean training data, the choice of the VAE decoder is still crucial. For good reconstruction 
from the latent space, we introduce a two-sided decoder, which captures the information flow from the input to the output node and vice versa. 
Specifically, nodes generated in the forward pass that are not connected to the output are connected to their predecessor in the backward decoder, and vice versa. \textcolor{black}{The union of both forward and backward decoded graphs will therefore likely contain all missing edges from both single decoders.
} 
D-VAE~\cite{zhang2019d} overcomes this problem of possible trailing nodes by incorporating a heuristic ``post-processing" step employing prior knowledge on the search space: It connects each non-output node with out-degree zero to the output node. 

\begin{figure*}[ht]
\centering
    \includegraphics[width=\textwidth, height=3cm]{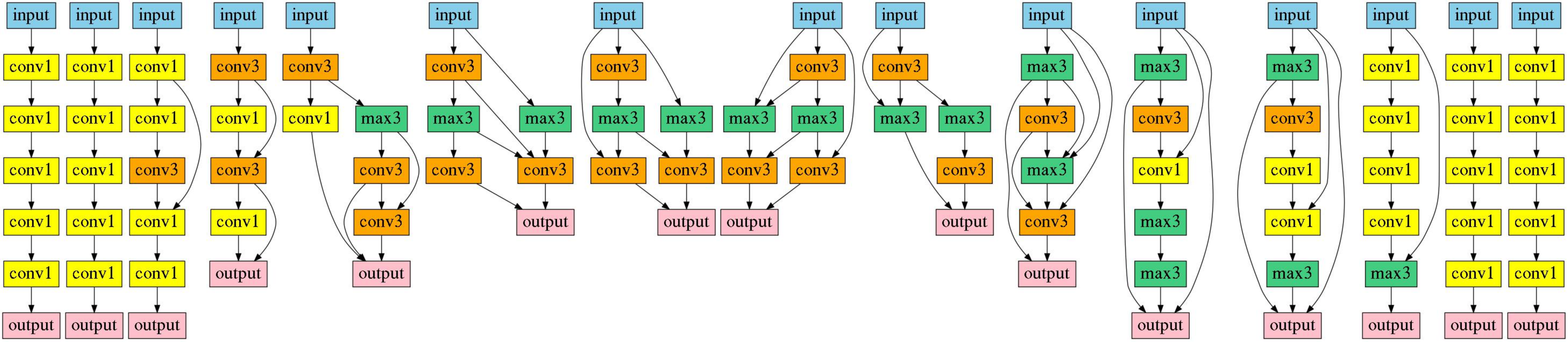} 
   \caption{Circle in the latent space with 14 equidistant sampled points on a hypersphere. 
   The graphs vary smoothly with increasing distance from the start.}
\label{fig:circle}
\end{figure*}

\section{Experiments} \label{sec:experiments}

We evaluate the proposed SVGe on three different, commonly used search spaces from the NAS literature.
%
\paragraph{NAS-Bench-101}
NAS-Bench-101 \cite{ying2019bench} is a tabular benchmark that consists of a cell-structured search space containing $423$k unique architectures evaluated for $4, 12, 36$ and $108$ epochs on the CIFAR-10 image classification task. The cell structure is limited to a number of nodes $\vert V \vert \leq 7$ (including the input and output node) and edges $\vert E \vert \leq 9$. The nodes represent an operation from the operation set $\mathcal{O} = \{1\times 1 ~\mathrm{conv.},\ 3 \times 3 ~\mathrm{conv.},\ 3 \times 3~ \mathrm{max~pooling}\}$. In our experiments, we use $90\%$ of the $423$k $(\mathrm{architecture},\ \mathrm{accuracy})$ pairs as training examples and $10\%$ as validation ones. 
\paragraph{ENAS Search Space}
The ENAS \cite{Pham2018} search space consists of architectures represented by a DAG with $\vert V \vert =8$ nodes (including the input and output node) and 6 operation choices on each of the non-input and non-output nodes. As \cite{zhang2019d}, we sample $19,020$ architectures from this space. 
%
\paragraph{NAS-Bench-201} 
NAS-Bench-201 \cite{dong2020} is a tabular benchmark, employing a cell-structured search space, with $15,625$ unique, sampled architectures trained and evaluated on CIFAR-10, CIFAR-100 and ImageNet-16-120 \cite{ImageNet16_120} for image classification. NAS-Bench-201 differs from NAS-Bench-101 in the cell representation: the nodes represent the sum of feature maps and the edges represent an operation from $\mathcal{O} = \{1\times 1 ~\mathrm{conv.},\ 3\times 3~\mathrm{conv.}, 3\times 3~\mathrm{avg~pooling}, \mathrm{skip}, \mathrm{zero}\}$. The DAG representing a NAS-Bench-201 architecture has $\vert V \vert =4 $ nodes and $\vert E \vert =6$ edges. 

While the NAS-Bench-101 benchmark provides the true performance of all fully trained architectures, ENAS does not provide any such value. Therefore, we use the weights of the optimized one-shot model as a proxy for the validation/test performance of the sampled architectures during optimization. We split the $(\mathrm{architecture}, \mathrm{accuracy})$ pairs into $90\%$ training and $10\%$ test examples. On NAS-Bench-201, we evaluate abilities of the proposed autoencoder and show the transferability of our BO and extrapolation results.
%

After testing its autoencoder abilities, we evaluate SVGe on performance prediction, Bayesian optimization and search space extrapolation.
In all our experiments, we set $\mathbf{h} \in \mathbb{R}^{250}$ for the concatenated node dimension and $\mathbf{h}_G \in \mathbb{R}^{56}$ for the latent space dimension. 
All the algorithms and routines are implemented using PyTorch~\cite{paszke2017automatic} and PyTorch Geometric~\cite{fey2019fast}\footnote{We provide our implementation at 
\url{https://github.com/jovitalukasik/SVGe}}.

\begin{figure}[t]
\centering
    \includegraphics[height=3.7cm, width=4.3cm]{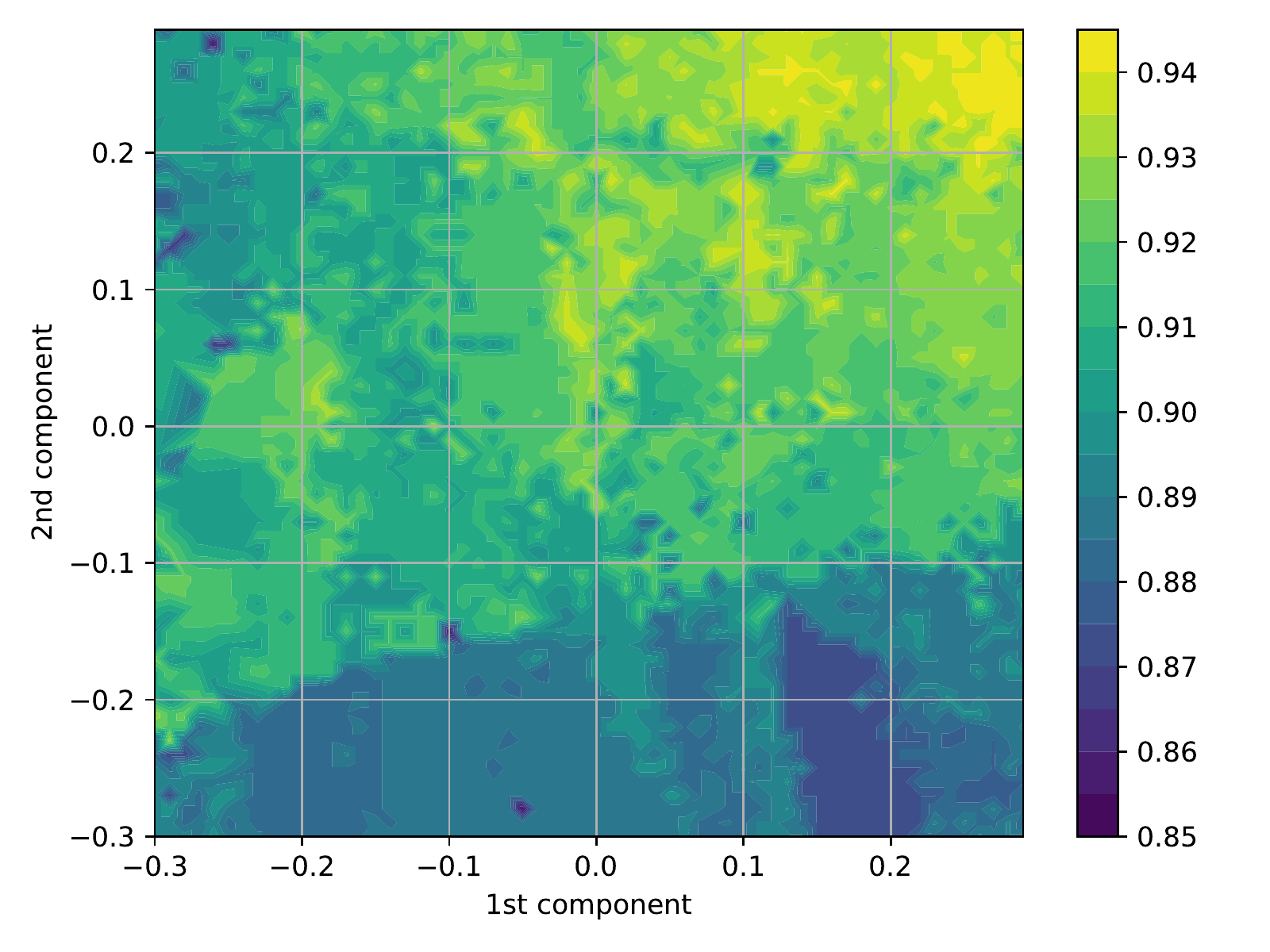}   
     \includegraphics[height=3.7cm]{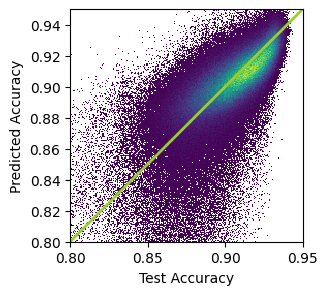}
   \caption{(left) Visualization of the first two principal components of the latent space for SVGe. (right) Performance prediction of the fine-tuned SVGe on the NAS-Bench-101 test data with a validation accuracy above $80 \%$}
\label{fig:grid} \label{fig:pp_NAS_density}
\end{figure}
\begin{table}[t]
\caption{VAE abilites on ENAS, NAS-Bench-101 and NAS-Bench-201 in $\%$. 
}
\centering
\small{
\begin{tabular}{@{\,}l@{\,\,\,}|@{\,\,\,}l@{\,\,\,}| @{\,\,\,}c@{\,\,\,}c@{\,\,\,}c@{\,\,\,}c@{\,}}
\toprule
 & Method & Accuracy & Validity & Uniqueness & Novelty  \\
\midrule
\multirow{3}{*}{\rotatebox{90}{ENAS}} &D-VAE \cite{zhang2019d} &$99.96$ & $100$& $ 37.26$ & $100$ \\
&DGMG \cite{li2018learning} & $99.29$ & $100$ & $37.55$ & $100$ \\
&SVGe &   $ 99.63$ &  $ 100$  &  $ 39.03$ & $100 $ \\
\midrule
\multirow{4}{*}{\rotatebox{90}{NB101}} &D-VAE  & $25.89$ & $82.55$ & $19.84$ & $16.52$ \\
& arch2vec \cite{Arch2vec} & $98.84$ & $43.70$ & $10$ & $82.84$\\
&DGMG  & $99.99$ &$89.7$ &$29.24$ &$16.72$ \\
& SVGe &$99.57$ & $79.16$ & $32.1$ & $16.37$ \\
\midrule
\multirow{3}{*}{\rotatebox{90}{NB201}} & arch2vec  & $99.99$ & $95.93$ & $7.33$ & $13.26$ \\
& DGMG  &$99.97$  & $100$ & $5.35$ & $12.62$ \\
& SVGe & $99.99$ & $100$ & $8.28$ & $10.24$ \\
\bottomrule
\end{tabular}%
}
 \label{table:gae_ability}
\end{table}
\subsection{Autoencoder Abilities.}\label{sec:exp_abilities} Following previous work \cite{zhang2019d, Jin2018}, we evaluate SVGe by means of \textit{reconstruction accuracy}, \textit{validity}, \textit{uniqueness} and \textit{novelty}. 
We evaluate these abilites on the ENAS, NAS-Bench-101 and NAS-Bench-201 search spaces and compare to \cite{zhang2019d} and \cite{Arch2vec}. As an ablation on ENAS, NAS-Bench-101 and NAS-Bench-201, we also adapted the generative model from \cite{li2018learning}, \textit{DGMG}, as a VAE with a single decoder to assess our encoder architecture. 
We train the models on $90 \%$ of the dataset and test it on the $10\%$ held-out data.

Table \ref{table:gae_ability} shows the results.
D-VAE~\cite{zhang2019d} and all our model variants show reasonable performance w.r.t. the \textit{reconstruction accuracy},  on ENAS
. On NAS-Bench-101 and 201, our approaches perform equally well and are comparable to arch2vec~\cite{Arch2vec}, while D-VAE performs poorly on NAS-Bench-101 and diverges on NAS-Bench-201. We hypothesize that D-VAE's hard constraints on the graph decoding are not suitable for NAS-Bench-101 and NAS-Bench-201. The resulting latent space cannot capture all relevant information. 

The \textit{validity} measures how many of the decoded samples are valid DAGs. Again, we see good overall performance on ENAS. On NAS-Bench-101 SVGe, DGMG and D-VAE perform comparably, 
while the validity for arch2vec is low, indicating that their decoder is not suited for graph generation. This trend is less severe yet observable on NAS-Bench-201. 

The \textit{uniqueness} ability measures the unique share of the valid decoded graphs. We argue that it can be seen as a measure for the latent space smoothness, which is particularly important for any kind of extrapolation from the training data. If the uniqueness is small, distinct and potentially distant latent points 
are decoded to the same output. This indicates that the latent space is heavily folded and non-smooth. 
While all approaches could be improved w.r.t. uniqueness, SVGe 
performs better than previous models. 
 
 The \textit{novelty} indicates the portion of graphs from the valid graph set which have not been observed during training. Here, values on ENAS are high and non-informative since only a small portion of the search space is covered by training data. On the two NAS-Bench variants, D-VAE and our models perform on par while the numbers for arch2vec are higher. The direct comparison of these values is impaired since arch2vec issues an overall lower number of valid graphs. 

We conclude that SVGe shows the best trade-off between accuracy, validity and uniqueness over all three search spaces.

\begin{table}[t]
\centering
\caption{Comparison of MSE and standard deviation for performance prediction on NAS-Bench-101.
}
\small
\begin{tabular}{@{}l@{\,\,}|@{\,\,}c@{\,\,\,}c@{\,\,\,}c@{\,\,}}
\toprule
 & \multicolumn{3}{c}{Performance Prediction}\\
 & {$1,000$} & {$10,000$} & {$100,000$}  \\ \midrule
Assess.\cite{Tang_Assessor}\hspace{0.cm}  & $0.0031$ {$\pm 0.0003$}  & $0.0026 ${$\pm 0.0002$}  & $ \mathbf{0.0016}${$\pm 0.0002$}   \\
D-VAE \cite{zhang2019d} \hspace{0.cm}  & $0.0039$ {$\pm 0.0003$}  & $0.0026 ${$\pm 0.0002$}  & $0.002${$\pm 9e-5$} \\
DGMG \cite{li2018learning} \hspace{0.cm}  & $0.0037$ {$\pm 0.0001$}  & $0.0027${$\pm 3 \mathrm{e}-6$} & $0.002 {\pm 0.0001}$\\
SVGe  & $\mathbf{0.0028}${$\pm 0.00002$}    & $\mathbf{0.0023}${$\pm 0.00004$}  &$0.002 ${$\pm 0.00003$}  \\ 
\bottomrule
\end{tabular}%
\label{tab:ass_rmse}
\end{table}
\subsection{Latent Space Smoothness Observations}
In Fig.~\ref{fig:grid} and \ref{fig:circle}, we visualize the smoothness of the SVGe graph embedding in the NAS-Bench-101 search space. Fig.~\ref{fig:grid} (left) visualizes the SVGe embedding. We plot equidistant points within a $[-0.3, 0.3]$ grid, given a 2D subspace of our training data with a validation accuracy above $75\%$ spanned by the first two principal components. Architectures with similar accuracies are close to each other and high accuracy architectures form clusters.  
Fig.~\ref{fig:circle} shows a unit circle in a randomly chosen orthogonal direction in the SVGe embedding space. We start from a flat net encoding in the latent space and randomly pick $14$ equidistant datapoints along the hypersphere returning to the start point. These datapoints are decoded and visualized as architectures. As one can see they change smoothly with changing only few operations and edges at each step. 

\subsection{Performance Prediction from Latent Space}\label{subsec:perfomance_prediction}
Next, we evaluate SVGe in terms of performance prediction on NAS-Bench-101 architectures. This allows for direct comparison to the recent work \cite{Tang_Assessor}. 
We train SVGe on $90\%$ of all $423$k datapoints in NAS-Bench-101 for reconstruction to obtain the latent space.
Then, we fine-tune the unsupervisedly trained model for performance prediction using a regressor, which is a four-layer MLP with ReLU non-linearities.
The SVGe model and the regressor are trained jointly for performance prediction on $1/ 10/ 100$k randomly sampled architectures with test accuracies from NAS-Bench-101. For comparison we also train D-VAE and DGMG in the same setting. 
We compare the ability to predict performances accurately on the validation set. Table~\ref{tab:ass_rmse} shows the MSE, which denotes the empirical squared loss between the predicted and ground truth data, and the standard deviation of $3$ runs.

Our proposed SVGe has a slightly lower MSE compared to 
\cite{Tang_Assessor}, which focuses precisely on this subproblem, when few annotated datapoints are given. This is important in particular for NAS, since every training sample corresponds to a fully evaluated architecture and is thus expensive. D-VAE and DGMG show high MSE for this small amount of training data. We expected this behavior for D-VAE because it already showed poor abilities in Sec.~\ref{sec:exp_abilities} on NAS-Bench-101. The low prediction accuracy for DGMG hints to potential overfitting in the autoencoder.
In  Fig.~\ref{fig:pp_NAS_density} (right), we plot the performance prediction ability of our model trained on $1$k sampled architectures from Table~\ref{tab:ass_rmse} for high-performing architectures (above $80\%$ test accuracy). This shows a strong correlation between predicted and true accuracies. 

\begin{table}[t]
\centering
\caption{BO on the ENAS and NAS-Bench-101 search space. SVGe slightly outperforms D-VAE~\cite{zhang2019d} on ENAS and reduces the runtime (in GPU hours) 
by a factor of~3.}
\small{
\begin{tabular}{@{\,}l@{\,\,}|@{\,\,}l@{\,\,}|c@{\,\,}c@{\,\,}c@{\,}}
\toprule
          & Method       & Test Acc.(\%)  &  Val Acc.(\%) &Runtime  \\ \midrule
\multirow{2}{*}{ENAS}              & D-VAE \cite{zhang2019d}&  $94.80 $  & - & 16  \\    
                                    & SVGe         & $\mathbf{95.11}$ &-& \textbf{5}      \\
\midrule
\multirow{3}{*}{NB101} &
\textit{oracle}&94.09&95.15&-\\
& DGMG \cite{li2018learning}  & $93.51$ & $94.08$ &-  \\
& SVGe  & $93.88$ & $94.60$ &-  \\
\bottomrule
\end{tabular}%
}
 \label{table:BO}\label{table:NAS-BO}
\end{table}


\subsection{Bayesian Optimization}\label{sec:BO}
We have seen in the previous experiments that the proposed SVGe generates a latent space which enables to interpolate from seen labels/performances and outperforms D-VAE and DGMG significantly. Next, we perform NAS via BO in the ENAS search space, in order to allow a fair comparison to D-VAE~\cite{zhang2019d} by exchanging only the D-VAE generative model with our SVGe and using exactly the same setup as in \cite{zhang2019d}. 
%
We perform $10$ iterations of batch BO (with a batch size of $50$) and average the results across $10$ trials based on a Sparse Gaussian Process  (SGP)~\cite{snelson} with $500$ inducing points and expected improvement \cite{mockus74} as acquisition function. 

We select the best $15$ architectures w.r.t. their weight-sharing accuracies and fully train them from scratch on CIFAR-10. As shown in Table~\ref{table:BO}, SVGe's best found architecture achieves an accuracy of $95.11\%$,
which is $0.31$ percent points better than the best found architecture using the D-VAE embedding. Table \ref{table:BO} also reports compute times for model training on ENAS. It shows that SVGe can be trained more efficiently than D-VAE.

Additionally, we perform BO on the NAS-Bench-101 search space with our SVGe model and optimize on validation accuracies. We train the SGP initially on $1$k randomly sampled architectures in each trial. 
Because of its low performance prediction on NAS-Bench-101, D-VAE is expected to perform poorly in this setting. To assess our results, in Table \ref{table:NAS-BO}, we report as \textit{oracle} the best NAS-Bench-101 architecture in terms of validation accuracy and its test accuracy. 
BO on SVGe yields a model with $94.60 \%$ validation and $93.88 \%$ test accuracy, \textcolor{black}{improving over the best found architecture using DGMG in terms of both validation and test accuracy. When using all our training data ($90 \%$ of NAS-Bench-101) to train the SGP, SVGe's best found architecture achieves a validation accuracy of $94.67 \%$.} This architecture yields a test accuracy of $94.26 \%$ on NAS-Bench-101 which is higher than the test accuracy of the best NAS-Bench-101 architecture in terms of validation accuracy. Note that the best \textit{oracle} test accuracy would be 94.45\% (at only 94.87\% validation acc.). 

 Since the D-VAE training diverges on the NAS-Bench-201 search space, we can not conduct a direct comparison. Yan et al.~\cite{Arch2vec} perform BO in their latent space, using DNGO \cite{DNGO} instead of SGP, and define the current state-of-the-art. 
 DNGO is suited for 
low-dimensional embedding spaces while it performs less well 
on high dimensional spaces as ours. 
Conversely, using SGP on low-dimensional embedding spaces is sub-optimal. Therefore, a direct comparison to arch2vec in terms of BO should be taken with caution. 
Performing BO in the SVGe generated latent space yields a test accuracy of $93.38 \%$ on the CIFAR-10 image classification task. 
 In comparison arch2vec yields a mean test accuracy of $94.18 \%$, which only leaves a small gap. 
%
Thus, SVGe is able to find well-performing architectures in all three search spaces.

\begin{table}[t]
\centering
\caption{Architecture extrapolation experiments.
}
\small
\begin{tabular}{l|l|cc}
\toprule
Dataset                                        & Method       & Val Acc. (\%) &  Test Acc. (\%) \\ \midrule
NB101-7 & \emph{oracle}& $95.15$ & $94.09$\\ 
NB101-8 & SVGe         & $\mathbf{95.18} $    & $ \mathbf{94.92} $  \\
\midrule
\multirow{2}{*}{ENAS-14}              & D-VAE+BO \cite{zhang2019d} &  $\mathbf{96.12}$       & - \\    
                                    & SVGe         &    96.09&  -     \\
                                               \bottomrule
\end{tabular}
 \label{table:extrapolation}
\end{table}

\begin{table}[t]
\centering
\caption{Dataset Transfer Learning to ImageNet 16-120. }
\small{
\begin{tabular}{l@{\,\,}|@{\,\,}c@{\,\,}c@{\,\,}}
\toprule
Method       & Test Acc.(\%)  &  Val Acc.(\%)   \\ \midrule
NAS-Bench-201 (optimal) \cite{dong2020}&    $47.31$ & $46.77$ \\    
ResNet \cite{dong2020} & $43.63$ & $44.53$ \\
SVGe + BO (NB101-7)         & $\mathbf{56.83}$ &  $54.70$    \\
SVGe + Zero-Shot (NB1018)  & $55.53$ & $\mathbf{55.13}$    \\
\bottomrule
\end{tabular}%
}
\label{table:ImageNet}
\end{table}

\subsection{Extrapolation Ability}\label{subsec:extrapolation}
Finally, we show that our smooth embedding space enables to find better architectures than the ones mentioned above even without dedicated optimization approaches by simple extrapolation from the labeled dataset. 
We employ the ability of SVGe to predict neural architectures' performances on CIFAR-10 with more nodes and edges than seen at training time in both NAS-Bench-101 and ENAS search spaces.

On the NAS-Bench-101 search space, we generate graphs (cells) containing 8 nodes. Note that our SVGe model has never seen such architectures during training 
(NAS-Bench-101 is limited to cells with up to 7 nodes). To generate these new graphs, we pick the best performing graph from NAS-Bench-101 based on the validation accuracy and expand it to graphs with 8 nodes, maintaining the  upper triangular matrices structure ($1,384$ graphs in total).  
From these graphs, we select $5$ samples with the highest predicted validation accuracy using SVGe (see Sec.~\ref{subsec:perfomance_prediction}, trained on $1$k graphs). These models are trained from scratch on CIFAR-10 using the training pipeline from \cite{ying2019bench}. As shown in  Table~\ref{table:extrapolation}, the architectures found by extrapolating using our SVGe model achieve a top-1 validation accuracy of $95.18 \%$ and a test accuracy of $94.92\%$ for graphs of length 8, which improves over $0.83 \%$ in test accuracy over the best 7-nodes architecture test accuracy. 

On the ENAS search space, we evaluate SVGe on the macro architecture containing a total of $14$ nodes (layers, including the input and output node) compared to architectures with $8$ nodes used during the SVGe training. We further fine-tune the embedding space by sampling $1$k architectures from the training set and train the SVGe together with the performance predictor. Note that this performance predictor uses the weight-sharing accuracies as proxy for the true accuracy of the fully trained architectures.
We select top $5$ architectures based on the predicted validation performance and again fully train them on CIFAR-10, using the settings from \cite{zhang2019d}. 
As shown in Table~\ref{table:extrapolation}, the best found architecture in the ENAS search space achieves a validation accuracy of $96.09 \%$ which is close to the one found by  
D-VAE. 
Note, D-VAE~\cite{zhang2019d} used a Bayesian optimization approach as in Sec. \ref{sec:BO} to find this architecture, whereas SVGe can achieve similar results by direct extrapolation (aka zero-shot prediction). 

Last, we test the transferabilty of the architectures found by our model. 
For that purpose, we train the best found architecture in the BO experiment and the top 1 architecture found via extrapolation on ImageNet16-120 \cite{ImageNet16_120} in the training scheme from \cite{dong2020}. As shown in Table~\ref{table:ImageNet} both architectures improve over a comparably deep ResNet architecture \cite{resnet} and the best NAS-Bench-201 architecture by a significant margin.

\section{Conclusion}
In this paper, we proposed SVGe, a Smooth Variational Graph embedding model for NAS. We give empirical results on SVGe encoding abilities and show that it applies more easily to new search spaces than previous approaches~\cite{zhang2019d}. 
We present results on the NAS-Bench-101, NAS-Bench-201 and ENAS search spaces and show good results for performance prediction surrogate models and Bayesian optimization in the smooth embedding space. Furthermore, we demonstrate the extrapolation abilities of SVGe to larger unseen graphs to find high-performing architectures. Image dataset transfer experiments to ImageNet16-120 also show that the found high-performing architectures can improve over the performance of comparable architectures by a significant margin. 

{\small
\bibliographystyle{IEEEtran}
\bibliography{IEEEabrv,bib_short}
}

\end{document}